\documentclass[11pt,a4paper]{article}
\usepackage[hyperref]{emnlp-ijcnlp-2019}
\usepackage{times}
\usepackage{latexsym}
\usepackage{url}
\usepackage{graphicx}
\usepackage{amsmath}
\usepackage{enumerate}
\usepackage{amsfonts}
\usepackage{array}
\usepackage{subfigure}
\usepackage{amssymb}
\usepackage{multirow}
\usepackage{booktabs}
\usepackage{footmisc}
\usepackage{floatrow}
\usepackage{mdwlist}
\usepackage{paralist}
\usepackage[linesnumbered,ruled,vlined]{algorithm2e}

\aclfinalcopy 

\title{Learning to Learn and Predict: \\A Meta-Learning Approach for Multi-Label Classification}

\author{Jiawei Wu \and Wenhan Xiong \and William Yang Wang \\
Department of Computer Science\\
University of California, Santa Barbara\\
Santa Barbara, CA 93106 USA\\
{\tt \{jiawei\_wu,xwhan,william\}@cs.ucsb.edu}}

\date{}

\begin{document}
\maketitle
\begin{abstract}
Many tasks in natural language processing can be viewed as multi-label classification problems. However, most of the existing models are trained with the standard cross-entropy loss function and use a fixed prediction policy (e.g., a threshold of 0.5) for all the labels, which completely ignores the complexity and dependencies among different labels. In this paper, we propose a meta-learning method to capture these complex label dependencies. More specifically, our method utilizes a meta-learner to jointly learn the training policies and prediction policies for different labels. The training policies are then used to train the classifier with the cross-entropy loss function, and the prediction policies are further implemented for prediction. Experimental results on fine-grained entity typing and text classification demonstrate that our proposed method can obtain more accurate multi-label classification results.
\end{abstract}

\section{Introduction}
\label{intro}
Multi-label classification aims at learning to make predictions on instances that are associated with multiple labels simultaneously, whereas in a classic multi-class classification setting, typically one instance has only one label. Multi-label classification is a common learning paradigm in a large amount of real-world natural language processing (NLP) applications, such as fine-grained entity typing~\cite{ling2012fine,shimaoka2016neural,abhishek2017fine,xin2018improving} and text classification \cite{nam2014large,liu2017deep,chen2017doctag2vec,wu2018reinforced}.

\begin{figure}[t]
\centering
\small
\includegraphics[width=1.0\columnwidth]{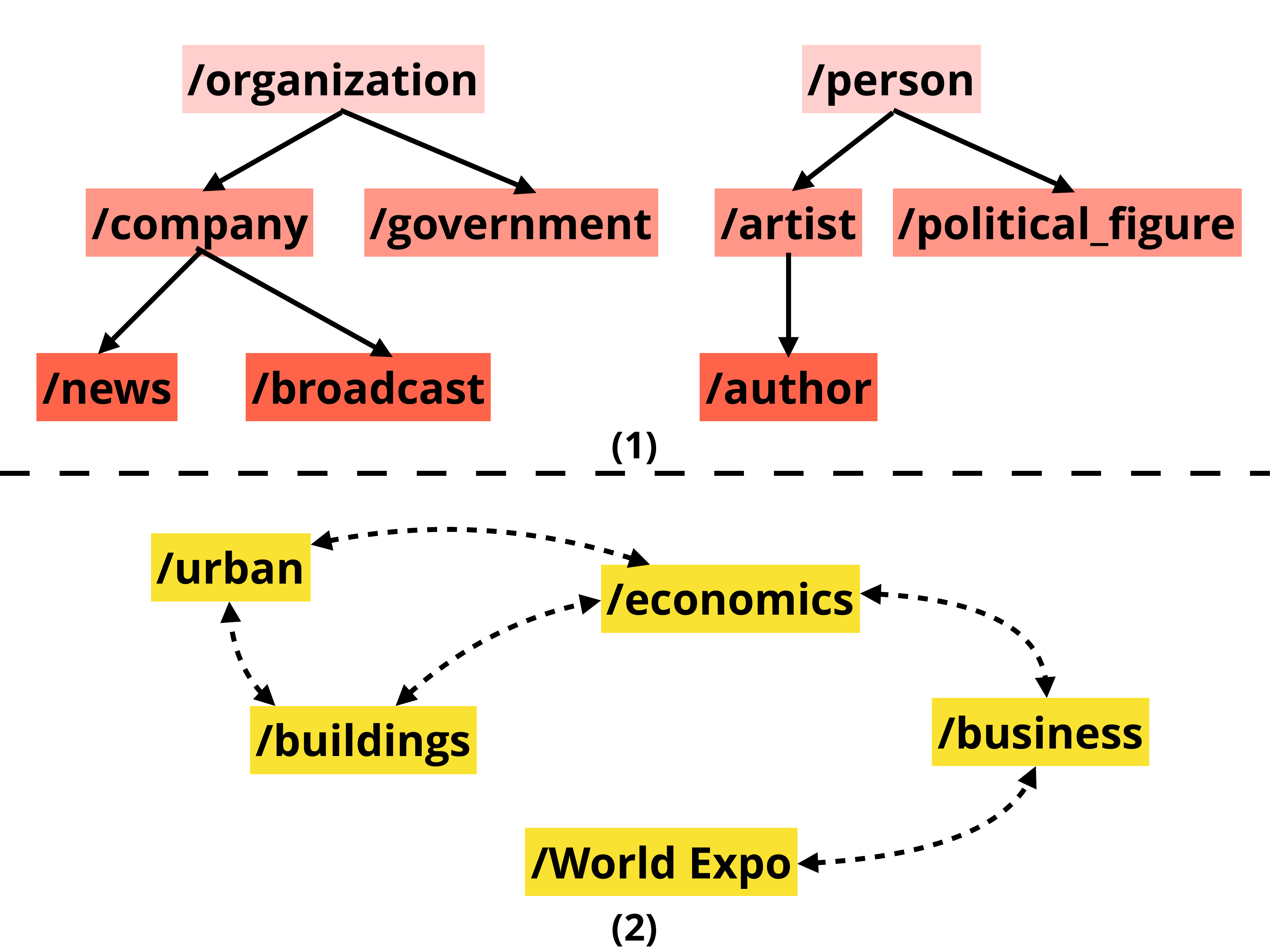}
\caption{Illustration of complex label dependencies in multi-label classification. (1) Labels are aligned with some knowledge graphs. In this case, labels are organized in a hierarchical dependency structure. (2) Even if labels do not demonstrate explicit correlation, they can still have implicit semantic dependencies.}\label{examples}
\end{figure}

Significant amounts of research studies have been dedicated to tackle the multi-label classification problem~\cite{zhang2014review}, from traditional statistical models \cite{zhang2007ml,zhou2012multi,surdeanu2012multi} to neural network-based models~\cite{nam2014large,shimaoka2016neural}. These models have variable structures, but usually, share the standard cross-entropy loss function for training. After training, these models tend to use a single prediction policy for all the labels to generate the final predictions. Actually, the above process is based on the assumption that there is no dependency among the labels. However, as shown in Figure \ref{examples}, this assumption is hard to be satisfied in real-world datasets, and the complex label dependencies receive little attention in multi-label classification \cite{dembszynski2010label,dembczynski2012label}. Owing to the impact of the label dependencies, although one object may have multiple labels simultaneously in the multi-label classification setting, the level of difficulty of prediction for different labels can vary a lot. Firstly, for those labels that are aligned with knowledge graphs, they usually indicate a hierarchical dependency structure \cite{ling2012fine,ren2016afet,ruder2016hierarchical}. It is intuitive that one trained classifier is easier to distinguish high-level parent labels, such as \texttt{/organization} and \texttt{/person}, but harder to distinguish low-level child labels, such as \texttt{/news} and \texttt{/broadcast}. Secondly, for those cases where labels do not demonstrate explicit correlation, the labels still contain implicit semantic dependencies, which is extremely common in the NLP field. For instance, the label \texttt{/urban} and \texttt{/economics} have obvious semantic correlation even if they are not organized in a hierarchical structure. Meanwhile, the labels with more implicit dependencies are easier to predict because they expose more semantic information during the training. These intuitions inspire our work on learning different training policies and prediction policies for different labels.

The training policies and prediction policies for all the labels can be viewed as a series of hyper-parameters. However, to learn high-quality policies, one needs to specify both explicit and implicit label dependencies, which is not manually realistic. To resolve both issues mentioned above, we propose a meta-learning framework to model these label dependencies and learn training and prediction policies automatically. Concretely, we introduce a joint learning formulation of the meta-learning method and multi-label classification. A gated recurrent unit (GRU)-based \cite{chung2014empirical} meta-learner is implemented to capture the label dependencies and learn these hyper-parameters during the training process of a classifier. Empirically, we show our method outperforms previous related methods on fine-grained entity typing and text classification problems. In summary, our contributions are three-fold:
\begin{itemize}
\item We are the first to propose a joint formulation of ``learning to learn'' and ``learning to predict'' in a multi-label classification setting.
\item Our learning method can learn a weight and a decision policy for each label, which can then be incorporated into training and prediction.
\item We show that our method is model-agnostic and can apply to different models in multi-label classification and outperform baselines.
\end{itemize}

In Section \ref{related}, we outline related work in multi-label classification and meta-learning. We then describe our proposed method in Section \ref{method}. We show experimental results in Section \ref{exp}. Finally, we conclude in Section \ref{conclusion}.

\section{Related Work}
\label{related}

\subsection{Multi-Label Classification}
Multi-label classification assigns instances with multiple labels simultaneously~\cite{tsoumakas2006review}. \cite{shore1980axiomatic,de2005tutorial} introduce prediction policies that weight the training loss function with external knowledge. However, in real-world multi-label classification, it is hard to obtain knowledge to determine prediction policies. As for the prediction policies, \newcite{yang2001study} select the thresholds that achieve the best evaluation measure on a validation set, while \newcite{lewis2004rcv1} utilize a cross-validation method to determine the thresholds. \newcite{lipton2014optimal} propose an optimal decision rule to maximize F1 measure with thresholds. However, most of the previous methods can be viewed as post-processing because the prediction policies are computed after training. Meanwhile, the ability of prediction policies to help train the classifier is less explored.

Compared to previous related methods, we propose a principled approach that learns a training policy and a prediction policy for each label automatically with modeling label dependencies implicitly. Furthermore, the prediction policies are learned during the training process instead of the post-processing, which can also help to train a better classifier.

\begin{figure*}[t]
\centering
\includegraphics[width=0.80\textwidth]{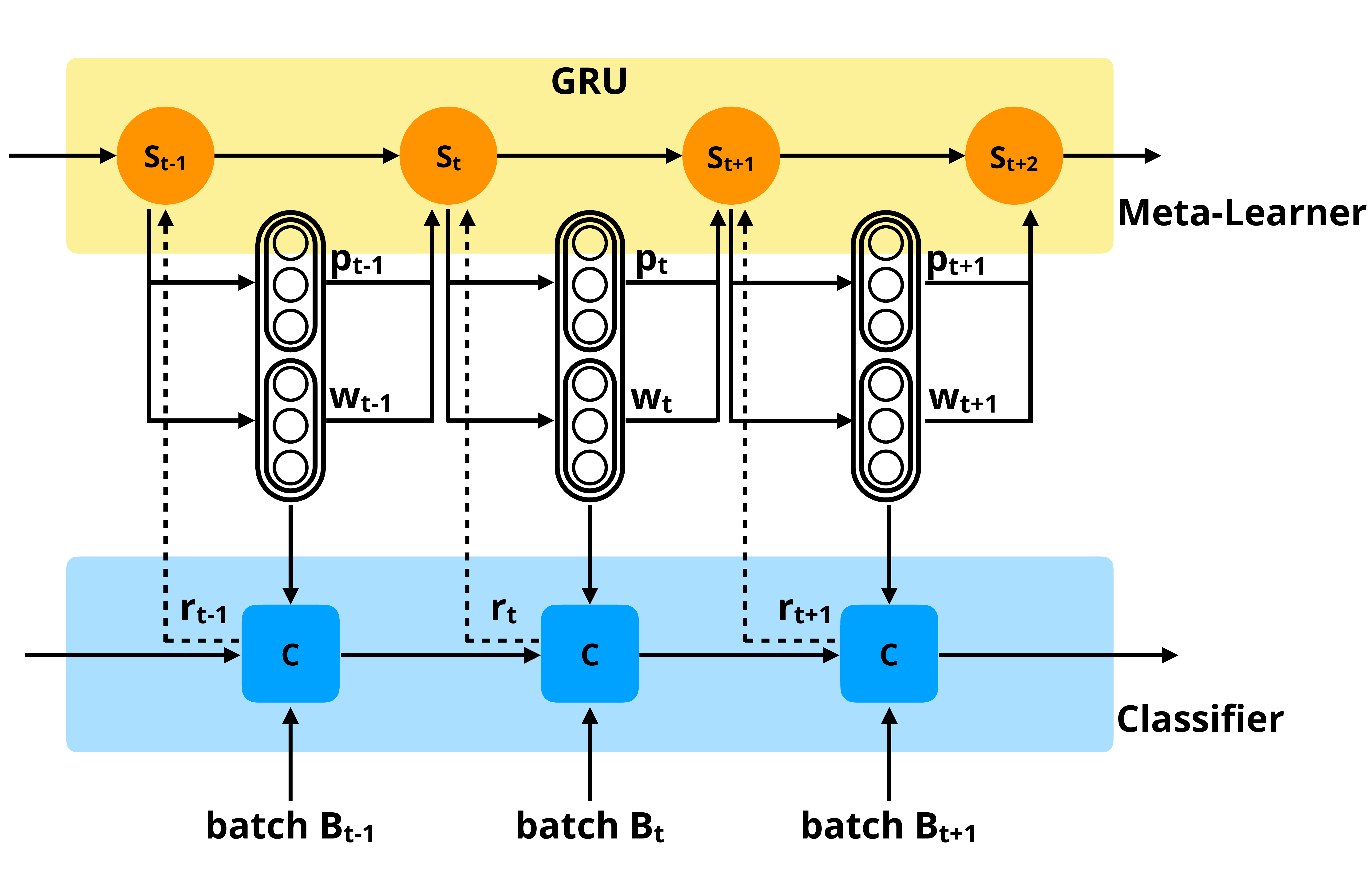}
\caption{The meta-learning framework for multi-label classification. \label{overview} }
\end{figure*}

\subsection{Meta-learning}
Meta-learning is a ``learning to learn'' method, in which a learner learns new tasks and another meta-learner learns to train the learner \cite{bengio1990learning,runarsson2000evolution,thrun2012learning}. There are two main types of meta-learning:
\begin{itemize}
\item learning a meta-policy to update model parameters~\cite{andrychowicz2016learning,mishra2017meta}.
\item learning a good parameter initialization for fast adaptation~\cite{duan2016rl,vinyals2016matching,finn2017model,snell2017prototypical,gu2018meta}.
\end{itemize}
In this paper, we propose to extend meta-learning algorithm for multi-label classification based on the first category. Instead of only training the model with meta-policy, we also consider prediction with meta-policy.

\section{Method}
\label{method}
In this section, we describe our meta-learning framework for learning and predicting multi-label classification in detail. In multi-label classification, the conventional methods usually learn classifiers with the standard cross-entropy loss function. After training, a single prediction policy (usually, a threshold of 0.5) is applied to all labels to generate the prediction. However, as mentioned in Figure \ref{examples}, these methods ignore the explicit and implicit label dependencies among the labels. To improve the performance of multi-label classification, our main idea is to learn high-quality training policies and prediction policies by meta-learning, which can also model the label dependencies.

We view the training policies and prediction policies as a series of hyper-parameters, and then formulate the training of the classifier as a sequential process. At each iteration (time step $t$), the classifier updates its parameters with one sampled batch. The exploration of this classifier's hyper-parameters can be modeled by a meta-learner. We first briefly introduce the classifier model used in multi-label classification. Then we describe the meta-learner, and the interaction between the meta-learner and the classifier. Finally, we explain how to train the meta-learner in our unified framework.

\subsection{Classifier Model}
\label{subsec:classifier}
As mentioned in Section \ref{intro}, many different model structures have already been explored in multi-label classification. Our framework is model-agnostic, which means different model structures are compatible if only they choose the cross-entropy loss function as the objective function. The classifier is represented as $C$. For a $N$- class multi-label classification, we represent the training policy as the vector $w = (w^{(1)}, w^{(2)}, \cdots, w^{(N)})$ and the prediction policy as the vector $p = (p^{(1)}, p^{(2)}, \cdots, p^{(N)})$, where $w^{(i)}$ and $p^{(i)}$ are the training weight and predicting threshold for the $i$-th class. $w_t$ and $p_t$ refer the weight vector and threshold vector at time step $t$. Then the goal of our framework is to learn a high-quality $w$ and $p$ for a certain classifier $C$.

To update the parameters of the classifier $C$, at each time step $t$, we sample a batch $B_t$ from the training set $U$. We then set a weighted cross-entropy objective function to update the $C$, which is defined as:
\begin{align}
\label{weighted_loss}
& L(\theta_t^C) = - \sum_i^{B_t}\sum_j^N \nonumber \\
& w_t^{(j)}N\{y^{*(j)}_i\log y^{(j)}_i + (1-y^{*(j)}_i)\log (1-y^{(j)}_i)\},
\end{align}
where $y_i^*$ indicates the ground truth prediction of the $i$-th sample from the $B_t$, and $y_i^{(j)}$ is the $j$-th entry of the corresponding output vector $y_i$. The standard cross-entropy loss function is a special case when $w_t^{(j)} = \frac{1}{N} (j = 1, 2, \cdots, N)$.

\subsection{Meta-Learner}
Meta-learning is a widely used reinforcement learning method to learn the meta-information of a machine learning system \cite{bengio1990learning,runarsson2000evolution,thrun2012learning}. The core of our proposed method is a meta-learner, which is trained to learn a training and a prediction policy for multi-label classification. At each time step $t$, the meta-learner observes the current state $s_t$, and then generate a training policy $w_t$ and a prediction policy $p_t$. Based on the policies $w_t$ and $p_t$, the parameters of classifier $C$ can be update with the sampled batch $B_t$ as described in Section \ref{subsec:classifier}. After training, the meta-learner receives a reward $r_t$. The goal of our meta-learner at each time step $t$ is to choose the two policies $w_t$ and $p_t$ that can maximize the future reward
\begin{equation}
R_t = \sum_{t'=t}^{T}r_{t'},
\end{equation}
where a training episode terminates at time $T$.

\subsubsection{State Representation}
The state representation, in our framework, is designed to connect the policy generation and the environment. At each time step $t$, the training policy, and the prediction policy are generated based on the state $s_t$. A reward will be computed based on the change of the environment. In our case, the performance change of the classifier $C$. In order to successfully explore the policy space and generate high-quality policies, the meta-learner needs to remember what similar policies have already been tried and make further generation based on these memories.

Based on the above intuition, we formulate the meta-learner as a recurrent neural network (RNN)-based structure. To simplify our method, we use a GRU in our experiments. The state representation $s_t$ is directly defined as the hidden state $h_t$ of the GRU at time step $t$. The $s_t$ is computed according to:
\begin{equation}
s_t = \text{GRU}(s_{t-1}, \begin{bmatrix}
p_{t-1} \\
w_{t-1} \\
\end{bmatrix}),
\end{equation}
where the input of GRU at time step $t$ is the concatenation of the prediction policy and training policy generated at time step $t-1$.

\subsubsection{Policy Generation}
\label{subsubsec:policy}
At each time step $t$, the meta-learner can generate two policies, the training policy $w_t$ and the prediction policy $p_t$. As mentioned in Section \ref{subsec:classifier}, both $w_t$ and $p_t$ are represented as a $N$-dimensional vector format.

To incorporate the training policy $w_t$ into the cross-entropy objective function in Equation \ref{weighted_loss} and keep the training gradients of the classifier in the same magnitude during the whole training episode, the condition $\sum_i^Nw_t^{(i)}=1$ for $w_t$ must be satisfied. Thus, at each time step $t$, the training policy is generated as:
\begin{equation}
w_t = \text{softmax}(W_ws_t+b_w).
\end{equation}

As for the prediction policy, it is obvious that $p_t^{(i)}\in(0,1)$ must be satisfied for $i=1, 2, \cdots, N$. Then we define the prediction policy as:

\begin{equation}
p_t = \text{sigmoid}(W_ps_t+b_p).
\end{equation}

The $s_t$ is the state representation of the meta-learner at time step $t$. $W_w$, $b_w$, $W_p$ and $b_p$ all are learnable parameters.

\begin{figure}[t]
\small
\centering
\includegraphics[width=0.90\columnwidth]{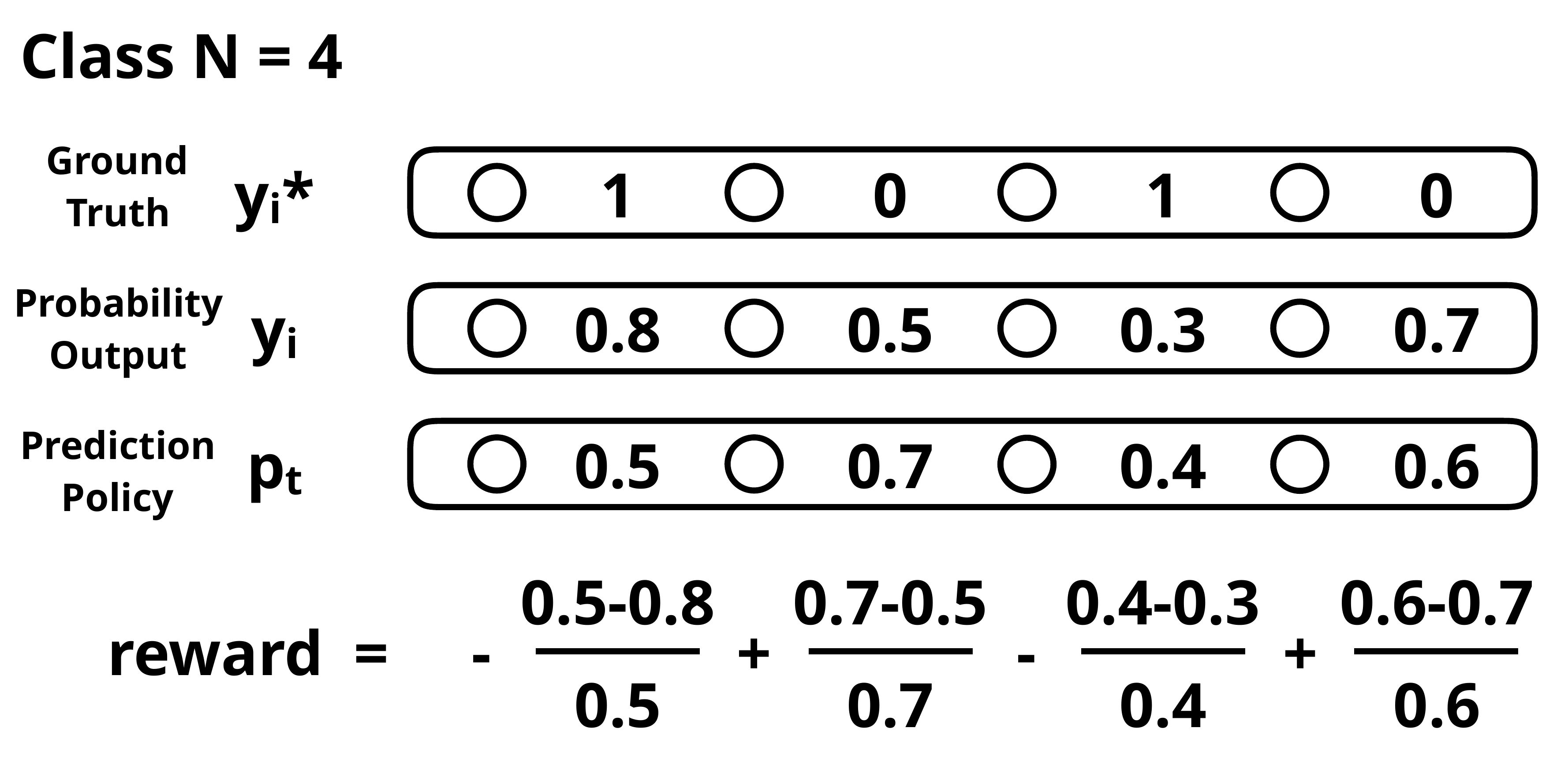}
\caption{A example about the computation process of reward (one sample with class $N=4$).}\label{reward_example}
\end{figure}

\subsubsection{Reward Function}
The meta-leaner is trained to generate high-quality training and prediction policies jointly to improve the performance of the classifier $C$. To capture this performance change, we design a reward function based on the probability distributions of samples.

At each time step $t$, we first generate the training policy $w_t$ and the prediction policy $p_t$ according to Section \ref{subsubsec:policy}. Then a batch $B_t$ is sampled from the training set and used to update the classifier $C$ based on Equation \ref{weighted_loss}. We evaluate the output probability distribution of all the samples from $B_t$ on the classifier $C$ and compute the reward as:
\begin{equation}
\label{compute_reward}
r_t = \sum_i^{B_t}\sum_{j=1}^N(-1)^{y_i^{*(j)}}\frac{p_t^{(j)}-y_i^{(j)}}{p_t^{(j)}},
\end{equation}
where $y_i$ is the output probability vector of the $i$-th sample from the batch $B_t$ and $y_i^*$ is the corresponding ground truth vector. The superscript $(j)$ represents the $j$-th entry of a vector. A simple example about how the reward is computed is shown in Figure \ref{reward_example}.

\subsection{Training and Testing}
The $\theta_{meta}$ is the set of all the parameters in the meta-learner, and the parameters can be trained by maximizing the total expect reward. The expected reward for an episode terminating at time step $T$ is defined as:
\begin{equation}
J(\theta_{meta}) = \mathbb{E}_\pi\lbrack\sum_{t=1}^T r_t\rbrack.
\end{equation}

$J(\theta_{meta})$ can be maximized by directly applying gradient-based optimization methods. We optimize it using policy gradient descent and implement a reward baseline to lower the variance during the training~\cite{sutton1984temporal}. The details of the training process are shown in Algorithm \ref{training}. During the training, the reason why we use sample batches instead of the full training set at each time step is that we want to explore more policy space with diversity instead of iteratively fitting the whole training set.

At test time, we rerun the meta-learner and the classifier simultaneously for one episode. However, we use the whole training set as a batch at each time step. The generated policies $w_T$ and $p_T$ is chosen as the final policies. We then train a classifier with the $w_T$-weighted cross-entropy objective function and test it based on the prediction policy $p_T$.

\begin{algorithm}[t]
\caption{The algorithm of our meta-learning framework.}\label{training}
Given a set of labeled training data $U$\\
Given a untrained classifier $C$ \\
\For{episode $\leftarrow$ 1 \KwTo M}{
    Initialize $w_0 \leftarrow (\frac{1}{N}, \frac{1}{N}, \cdots, \frac{1}{N}) \in \mathbb{R}^N$  \\
    Initialize $p_0 \leftarrow (0.5, 0.5, \cdots, 0.5) \in  \mathbb{R}^N$ \\
    \For {time step $t$ $\leftarrow$ 1 \KwTo T}{
       $s_t \leftarrow \text{GRU}(s_{t-1}, \begin{bmatrix} p_{t-1} \\ 
       w_{t-1} \\ \end{bmatrix})$ \\
       $w_t \leftarrow \text{softmax}(W_w s_t + b_w)$ \\
       $p_t \leftarrow \text{sigmoid}(W_p s_t + b_p)$ \\
       
       Sample a batch $B_t$ from $U$ \\
       Update $C$ using $B_t$ with $w_t$-based objective function in Equation \ref{weighted_loss} \\
       Compute reward $r_t$ with $p_t$ in Equation \ref{compute_reward}
    }
    Update $\theta_{meta}$ using $g \propto \nabla_{\theta}J(\theta_{meta})$
}
\end{algorithm}

\section{Experimental Setups}
\label{exp}
We evaluate our proposed method in following two settings: (1) \textbf{Fine-grained Entity Typing}, where the labels have explicit hierarchical dependencies; (2) \textbf{Text Classification}, where one needs to model the implicit label dependencies.

\subsection{Baselines}
Since our method is model-agnostic, we directly employ the state-of-the-art (SOTA) models for both two tasks. The details of SOTA models will be discussed in Section \ref{subsubsec:entity_sota} and \ref{subsubsec:text_sota}. We compare our method with multiple baselines:
\begin{itemize}
\item \textbf{Hierarchy-Aware Training Policy}: To explicitly add hierarchical information during the training, the labels are first given integer weights. For instance, the first-level (parent) labels are given weight $1$ while the third-level labels are given $3$. All the integer weights then are normalized to add into the cross-entropy loss function in Equation \ref{weighted_loss}.
\item \textbf{SCutFBR Prediction Policy}: The rank-based policy method RCut and proportion-based assignments PCut are jointly considered to set prediction policies for different labels after obtaining the trained classifier \cite{yang2001study}.
\item \textbf{ODR Prediction Policy}: After training, an optimal decision rule is to implement to choose prediction policies based on maximizing micro F1 scores \cite{lipton2014optimal}.
\item \textbf{Predictions-as-Features}: The model trains a classifier for each label, organize the classifiers in a partially ordered structure, and take predictions produced by the former classifiers as the latter classifiers' features \cite{li2015multi}.
\item \textbf{Subset Maximization}: The model views the multi-label prediction problem as classifier chains, and then replace classifier chains with recurrent neural networks \cite{nam2017maximizing}.
\end{itemize}

\section{Fine-grained Entity Typing}
\label{entity_typing}
Entity type classification is the task for assigning semantic types to entity mentions based on their context. For instance, the system needs label the entity \emph{``San Francisco Bay''} as \texttt{/location, /location/region} based on its context \emph{``... the rest of San Francisco Bay, a spot ...''}.

In a fine-grained classification setting, entities are aligned with knowledge graphs \cite{ling2012fine}, which typically makes labels be arranged in a hierarchical structure. We utilize our meta-learning framework to tackle this problem and evaluate our method.

\begin{table}[t]
\small
\begin{center}
\begin{tabular}{l|c|c|c}
\toprule 
\textbf{Datasets} & \textbf{FIGER} & \textbf{OntoNotes} & \textbf{BBN} \\
\midrule
\#Types & 128 & 89 & 47 \\
Max Hierarchy Depth & 2 & 3 & 2 \\
\#Training & 2,690,286 & 220,398 & 86,078 \\
\#Testing & 563 & 9,603 & 13,187 \\
\bottomrule
\end{tabular}
\end{center}
\caption{\label{stat_entity}The statistics of entity typing datasets.}
\end{table}

\paragraph{Datasets}
We evaluate our method on three widely-used fine-grained entity typing datasets, FIGER, OntoNotes, and BBN, which are pre-processed by \newcite{ren2016afet}. The statistics of these datasets are listed in Table \ref{stat_entity}.

\textbf{FIGER}: The training data is automatically generated by distant supervision, and then aligned with Freebase. The test data is collected from news reports and manually annotated by \newcite{ling2012fine}.

\textbf{OntoNotes}: The training sentences are collected from OntoNotes text corpus \cite{weischedel2013ontonotes}, and linked to Freebase. \newcite{gillick2014context} releases a manually annotated test dataset.

\textbf{BBN}: The dataset consists of sentences from Wall Street Journal articles, which is entirely manually annotated \cite{weischedel2005bbn}.

\begin{figure}[t]
\centering
\small
\includegraphics[width=0.90\columnwidth]{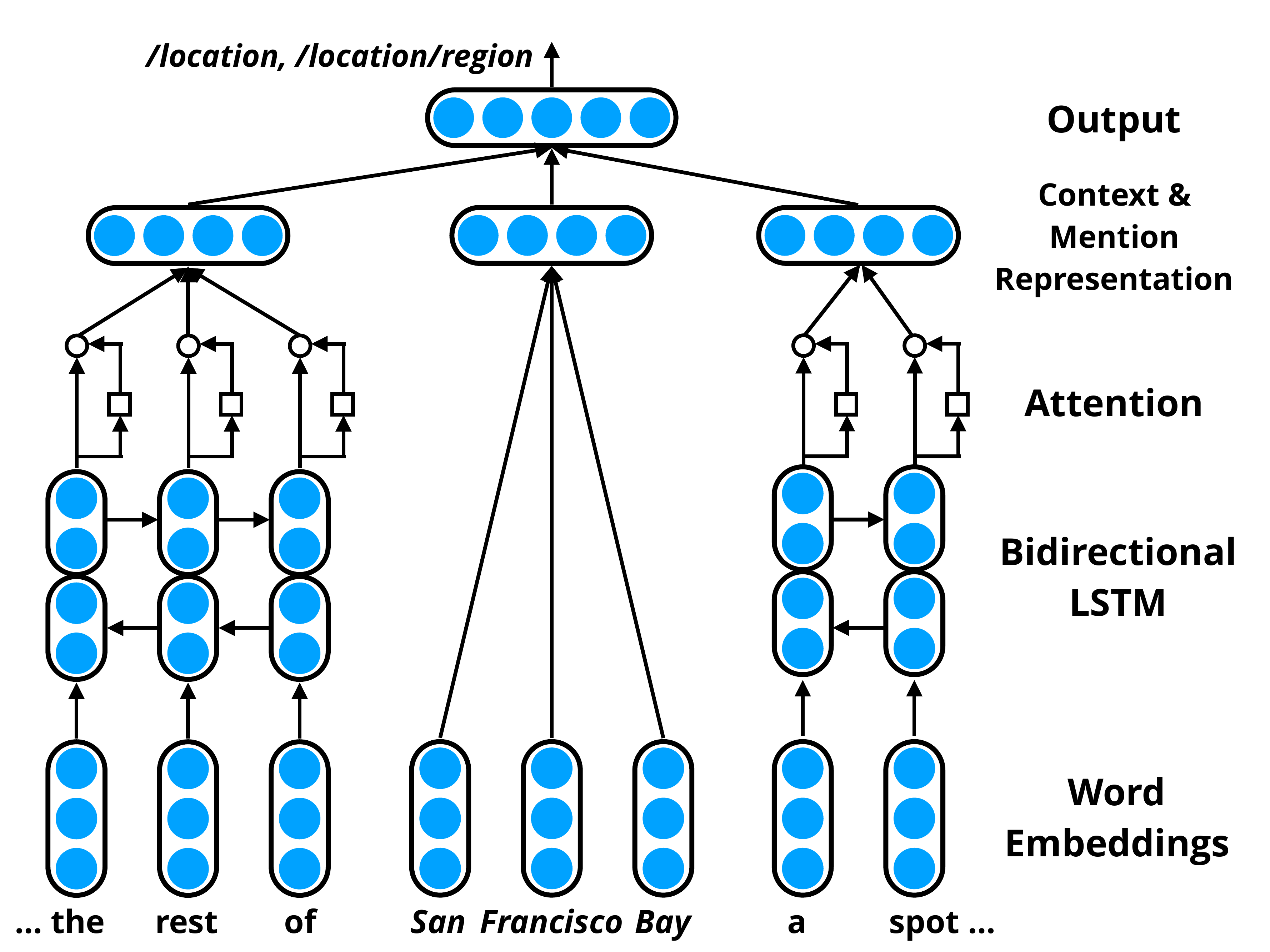}
\caption{The SOTA model structure for fine-grained entity type classification~\cite{shimaoka2016neural,abhishek2017fine}.}\label{bilstm}
\end{figure}

\paragraph{Implementation Details}
\label{subsubsec:entity_sota}
We choose the SOTA model of fine-grained entity typing as the classifier $C$ \cite{shimaoka2016neural,abhishek2017fine}. The overall model structure is shown in Figure \ref{bilstm}. Given en entity and its context sentence, the words are initialized with word embeddings \cite{mikolov2013distributed}. The mention representation is simply computed by averaging the word embeddings of entity mention words. As for the context representation, a bidirectional LSTM and attention mechanism are used to encode left and right context representation separately. The attention operation is computed using a Multi-Layer Perceptron (MLP) as follows:
\begin{equation}
a_i = \sigma \{ W_1 \text{tanh}(W_2
\begin{bmatrix}
\overrightarrow{h}_i \\
\overleftarrow{h}_i \\
\end{bmatrix}
)\},
\end{equation}
where $W_1$ and $W_2$ are parameter matrices of the MLP, and $\begin{bmatrix}
\overrightarrow{h}_i \\
\overleftarrow{h}_i \\
\end{bmatrix}$ is the hidden state of Bi-LSTM at the $i$-th position. Note that the parameters of MLP are shared by all the entities. The left, right context representations and mention representation are concatenated as the feature vector. A softmax layer is then implemented to perform final prediction. The standard cross-entropy loss function is used and the thresholds of all the labels are set as $0.5$. To avoid overfitting, we employ dropout operation on entity mention representation.

\begin{figure}[t]
\small
\centering
\includegraphics[width=0.80\columnwidth]{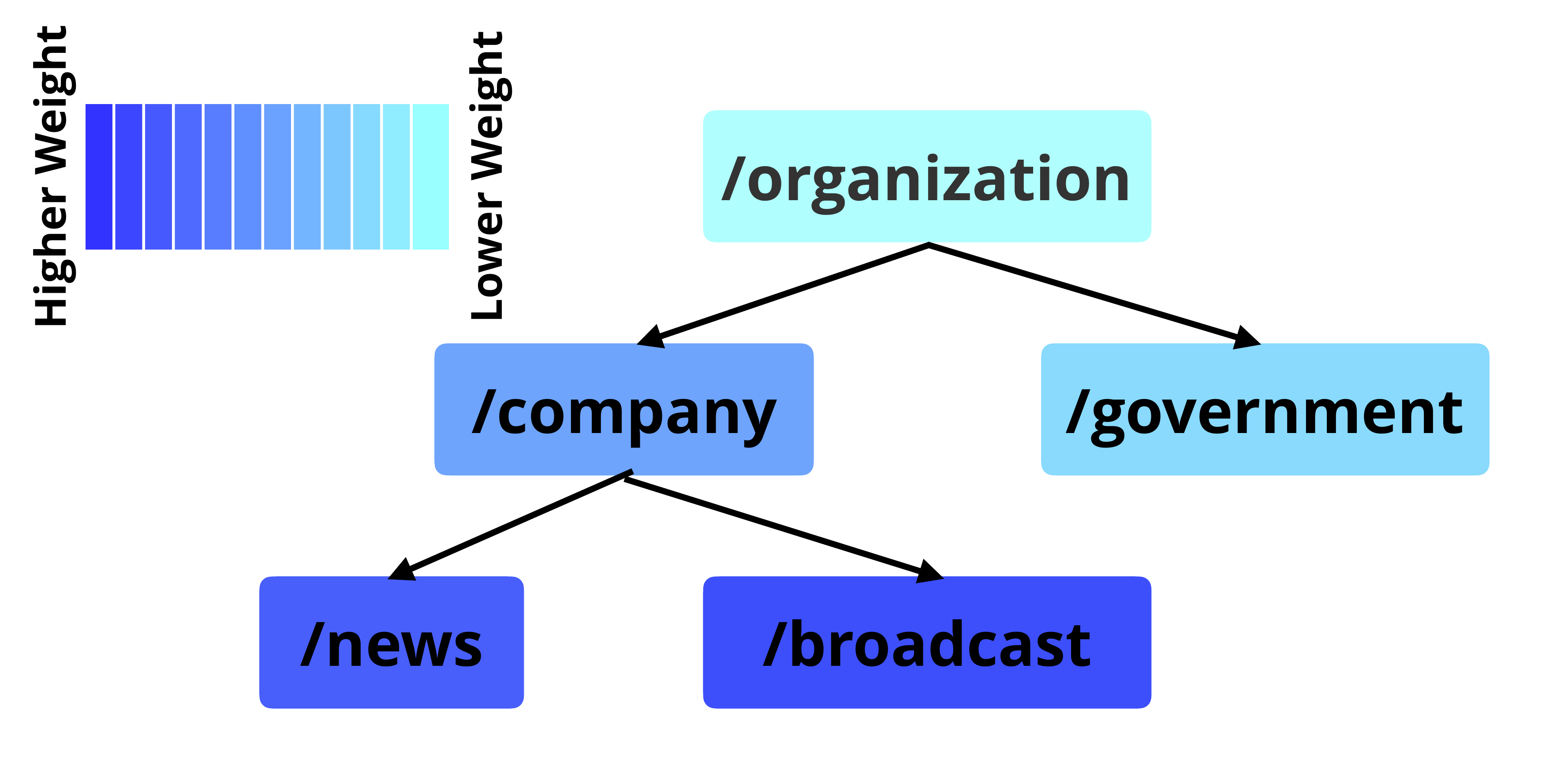}
\caption{A case study about weights. See Section \ref{subsubsec:weight_analysis} for the detailed explanation.}\label{weight_analysis}
\end{figure}

Following the previous researches \cite{ling2012fine,ren2016afet}, we use strict accuracy, loose macro, and loose micro scores to evaluate the model performance.

\begin{table*}[ht]
\small
\caption{\label{typing_results}The experimental results on fine-grained entity typing datasets. Acc.: Accuracy.}
\begin{center}
\begin{tabular}{@{}l|c|c|c|c|c|c|c|c|c@{}}
\toprule 
\textbf{Datasets} & \multicolumn{3}{@{}c|}{\textbf{FIGER}} & \multicolumn{3}{@{}c|}{\textbf{OntoNotes}} & \multicolumn{3}{@{}c@{}}{\textbf{BBN}} \\ 
\midrule
\textbf{Metrics} & \textbf{Acc.} & \textbf{Macro} & \textbf{Micro} & \textbf{Acc.} & \textbf{Macro} & \textbf{Micro} & \textbf{Acc.} & \textbf{Macro} & \textbf{Micro} \\
\midrule
SOTA & 0.659 & 0.807 & 0.770 & 0.521 & 0.686 & 0.626 & 0.655 & 0.729 & 0.751 \\
SOTA+Hier-Training & 0.661 & 0.812 & 0.773 & 0.532 & 0.690 & 0.640 & 0.657 & 0.735 & 0.754  \\
SOTA+Meta-Training & 0.670 & 0.817 & 0.779 & 0.539 & 0.702 & 0.648 & 0.662 & 0.736 & 0.761 \\
\midrule
SOTA+ScutFBR-Prediction & 0.662 & 0.814 & 0.782 & 0.542 & 0.695 & 0.650 & 0.661 & 0.736 & 0.758 \\
SOTA+ODR-Prediction & 0.669 & 0.818 & 0.782 & 0.537 & 0.703 & 0.648 & 0.664 & 0.738 & 0.764 \\
SOTA+Meta-Prediction & 0.674 & 0.823 & 0.786 & 0.544 & 0.709 & 0.657 & 0.671 & 0.744 & 0.769 \\
\midrule
Predictions-as-Features & 0.663 & 0.816 & 0.785 & 0.544 & 0.699 & 0.655 & 0.663 & 0.738 & 0.761 \\
Subset Maximization & 0.678 & 0.827 & 0.790 & 0.546 & 0.713 & \textbf{0.661} & 0.673 & 0.748 & 0.772 \\
\midrule
SOTA+Meta-Training-Prediction & \textbf{0.685} & \textbf{0.829} & \textbf{0.794} & \textbf{0.552} & \textbf{0.719} & \textbf{0.661} & \textbf{0.678} & \textbf{0.752} & \textbf{0.775} \\
\bottomrule
\end{tabular}
\end{center}
\end{table*}

\paragraph{Results}
The results of fine-grained entity typing are shown in Table \ref{typing_results}. From the results, we observe that: (1) Our meta-learning method can outperform all the baselines in all three metrics, which indicates the capability of our methods in improving the multi-label classification. (2) The SOTA model trained with learned training policy outperforms the one with hierarchy-aware training policy. It is because that our learned training policy can capture not only explicit hierarchical dependencies of labels, but also model other implicit label dependencies. (3) The three different prediction policies can improve the performance of the SOTA classifier. The results are consistent with previous researches that choosing a good prediction policy is an effective way to improve the performance \cite{fan2007study,lipton2014optimal}. (4) Compared with OntoNotes and BBN datasets, the FIGER show a relatively less improvement when applying these policies. The reason is that the test set of FIGER is not fine-grained enough (e.g., over 38\% of entities are only annotated with \texttt{/person} and no more fine-grained labels) \cite{xin2018improving}.

\paragraph{Algorithm Robustness}
\label{entity_analysis}
Previous researches \cite{morimoto2005robust,henderson2017deep} show that reinforcement learning-based methods usually lack robustness and are sensitive to the initialization, seeding datasets and pre-trained steps. Thus, we design an experiment to detect whether the trained meta-leaner is sensitive to the initialization. During the test time, instead of using the same initialization in Algorithm \ref{training}, we randomly initialize the $w_0$ and $p_0$ and learn $10$ groups of policies $w_T$ and $p_T$. For each group of policies, we train a classifier with $w_T$ and evaluate it with $p_T$ using the same metric. The results are shown in Table \ref{entity_robust}. The results demonstrate that our trained meta-learner is robust to different initialization, which indicates that the meta-learner in our method can generate high-quality and robust training and prediction policies to improve the multi-label classification.

\begin{table}[t]
\caption{\label{entity_robust}The robustness analysis on the FIGER dataset.}
\small
\begin{tabular}{l|c|c|c|c}
\toprule
\textbf{Metrics} & \textbf{Best} & \textbf{Worst} & \textbf{Average} & \textbf{STDEV} \\
\midrule
Accuracy & 0.689 & 0.679 & 0.681 & 0.0043 \\
Macro-F1 & 0.835 & 0.821 & 0.827 & 0.0039 \\
Micro-F1 & 0.796 & 0.787 & 0.789 & 0.0036 \\
\bottomrule
\end{tabular}
\end{table}

\paragraph{Weight Analysis}
\label{subsubsec:weight_analysis}
To analyze whether our meta-learner can really model the label dependencies, we perform a simple case study. In fine-grained entity classification, we choose $5$ labels that are originally organized in a hierarchical structure from the OntoNotes dataset. The corresponding entries of the training policy vector $w_T$ are extracted and expressed with colors. The case study is shown in Figure \ref{weight_analysis}. From the results, we can observe that high-level (parent) labels tend to have less training weights than low-level (child) labels. The results are consistent with our intuition that the labels with more dependencies expose more semantic information during the training.

\begin{table}[t]
\small
\begin{center}
\begin{tabular}{l|c|c}
\toprule
\textbf{Datasets} & \textbf{Reuters-21578} & \textbf{RCV1-V2} \\
\midrule
\#Labels & 90 & 103 \\
\#Average Labels/instance & 1.13 & 3.24 \\
\#Training & 7,769 & 781,265 \\
\#Testing & 3,019 & 23,149 \\
\bottomrule
\end{tabular}
\end{center}
\caption{\label{stat_text}The statistics of text classification datasets.}
\end{table}

\begin{table*}[ht]
\small
\caption{\label{text_results}The experimental results on text classification datasets.}
\begin{center}
\begin{tabular}{@{}l|c|c|c|c|c|c@{}}
\toprule 
\textbf{Datasets} & \multicolumn{3}{@{}c|}{\textbf{Reuters-21578}} & \multicolumn{3}{@{}c@{}}{\textbf{RCV1-V2}} \\ 
\midrule
\textbf{Metrics} & \textbf{Accuracy} & \textbf{Macro-F1} & \textbf{Micro-F1} & \textbf{Accuracy} & \textbf{Macro-F1} & \textbf{Micro-F1} \\
\midrule
CNN & 0.537 & 0.472 & 0.841 & 0.616 & 0.642 & 0.838  \\
CNN+Meta-Training & 0.542 & 0.476 & 0.843 & 0.631 & 0.655 & 0.852  \\
\midrule
CNN+ScutFBR-Prediction & 0.549 & 0.477 & 0.849 & 0.634 & 0.651 & 0.856  \\
CNN+ODR-Prediction & 0.541 & 0.475 & 0.848 & 0.630 & 0.653 & 0.849  \\
CNN+Meta-Prediction & 0.549 & 0.479 & 0.851 & 0.639 & 0.658 & 0.857  \\
\midrule
Predictions-as-Features & 0.539 & 0.476 & 0.845 & 0.621 & 0.644 & 0.847  \\
Subset Maximization & 0.543 & 0.478 & 0.849 & 0.632 & 0.660 & 0.859  \\
\midrule
CNN+Meta-Training-Prediction & \textbf{0.556} & \textbf{0.483} & \textbf{0.854} & \textbf{0.647} & \textbf{0.669} & \textbf{0.864} \\
\bottomrule
\end{tabular}
\end{center}
\end{table*}

\section{Text Classification}
\label{text_classifcation}
Text classification is a classic problem for NLP, where one needs to categorized documents into pre-defined classes \cite{nam2014large,liu2017deep,chen2017doctag2vec}. We choose the datasets in which samples have multiple labels and evaluate our model on text classification problem.

\begin{table}[t]
\caption{\label{text_robust}The robustness analysis on the RCV1 dataset.}
\small
\begin{tabular}{l|c|c|c|c}
\toprule 
\textbf{Metrics} & \textbf{Best} & \textbf{Worst} & \textbf{Average} & \textbf{STDEV} \\
\midrule
Accuracy & 0.652 & 0.641 & 0.646 & 0.0028 \\
Macro-F1 & 0.678 & 0.654 & 0.663 & 0.0041 \\
Micro-F1 & 0.874 & 0.855 & 0.863 & 0.0033 \\
\bottomrule
\end{tabular}
\end{table}

\paragraph{Datasets}
Following the settings in \cite{nam2014large}, we choose two multi-label datasets, Reuters-21578 and RCV1-V2, to test our method. The statistics of two datasets are listed in Table \ref{stat_text}.

\textbf{Reuters-21578}: The instances are collected Reuters news articles during the period 1987 to 1991. We use the same training/test split as previous work \cite{yang2001study,nam2014large}.

\textbf{RCV1-V2}: RCV1-V2 collects newswire stories from Reuters \cite{lewis2004rcv1}. The training and test dataset originally consist of $23,149$ train and $781,265$ test instances, but we switch them to better training and evaluation \cite{nam2014large}.

\paragraph{Setup}
\label{subsubsec:text_sota}
Many researches have proved convolutional neural networks (CNN) are effective in extracting information for text classification \cite{lecun1998gradient,kim2014convolutional,zhang2015character}. Following the \cite{kim2014convolutional}, we set a CNN model as the classifier $C$ to evaluate our method. Concretely, we use CNN-non-static mentioned in \newcite{kim2014convolutional}, which means we initialize the word embeddings with pre-trained Word2Vec and update the word embeddings during the training. The standard cross-entropy loss function is implemented, and the thresholds of all the classes are set as 0.5.

We still use strict accuracy, loose macro, and loose micro scores to evaluate the model performance following the settings in \cite{lewis2004rcv1,yang2012multilabel,nam2014large}.

\paragraph{Results}
The results of text classification are shown in Table \ref{text_results}. From the results, we can observe that: (1) Our meta-learning method can outperform all the baselines on two text classification tasks, which indicates that our approach is consistent with different tasks. (2) Compared with RCV1-V2, the classification results on Reuters-21578 show less improvement. The reason is that the number of average labels per instance in Reuters-21578 is 1.13, while the number is 3.24 for RCV1-V2 according to Table \ref{stat_text}. That means the multi-label classification on Reuters is close to the multi-class classification. There are little potential label dependencies in Reuters-21578.

\paragraph{Algorithm Robustness}
Similar to Section \ref{entity_analysis}, we evaluate whether our trained meta-learner is sensitive to the initialization. We follow the same steps mentioned in Section \ref{entity_analysis}, and show the results in Table \ref{text_robust}. The results indicate that the robustness of our meta-learner is consistent within different tasks and model structures, which again shows that the trained meta-learner can generate high-quality and robust policies.

\section{Conclusion}
\label{conclusion}
In this paper, we propose a novel meta-learner to improve the multi-label classification. By modeling the explicit and implicit label dependencies automatically, the meta-learner in our model can learn to generate high-quality training and prediction policies to help both the training and testing process of multi-label classifiers in a principled way. We evaluate our models on two tasks, fine-grained entity typing and text classification. Experimental results show that our method outperforms other baselines.

\section*{Acknowledgments}
The authors would like to thank the anonymous reviewers for their thoughtful comments. This research was supported in part by DARPA Grant D18AP00044 funded under the DARPA YFA program. The authors are solely responsible for the contents of the paper, and the opinions expressed in this publication do not reflect those of the funding agencies.

\bibliography{emnlp-ijcnlp-2019}
\bibliographystyle{acl_natbib}

\end{document}